\documentclass[10pt,twocolumn,letterpaper]{article}

\usepackage{cvpr}              

\usepackage{graphicx}
\usepackage{amsmath}
\usepackage{amssymb}
\usepackage{booktabs}
\usepackage{epstopdf}
\graphicspath{ {figures/} }

\usepackage[pagebackref,breaklinks,colorlinks]{hyperref}

\usepackage[capitalize]{cleveref}
\crefname{section}{Sec.}{Secs.}
\Crefname{section}{Section}{Sections}
\Crefname{table}{Table}{Tables}
\crefname{table}{Tab.}{Tabs.}

\def\cvprPaperID{42} 
\def\confName{CVPR}
\def\confYear{2022}

\begin{document}

\title{
Adversarial 
Machine Learning
Attacks Against Video Anomaly Detection Systems}

\author{Furkan Mumcu, Keval Doshi, Yasin Yilmaz\\
University of South Florida\\
4202 E Fowler Ave, Tampa, FL 33620\\
{\tt\small \{furkan, kevaldoshi, yasiny\}@usf.edu}
}
\maketitle

\begin{abstract}

Anomaly detection in videos is an important computer vision problem with various applications including automated video surveillance. 
Although 
adversarial attacks on image 
understanding models have been heavily investigated, there is not much work on adversarial machine learning targeting video understanding models 
and no previous work which focuses on video anomaly detection. 
To this end, we investigate an adversarial machine learning attack against video anomaly detection systems, that can be implemented via an easy-to-perform cyber-attack. 
Since 
surveillance cameras are usually 
connected to the server running the anomaly detection model through a wireless network,
they are prone to cyber-attacks targeting the wireless connection. 
We demonstrate how Wi-Fi deauthentication attack, a notoriously easy-to-perform and effective denial-of-service (DoS) attack, can be utilized to generate adversarial data for video anomaly detection systems. Specifically, we apply several effects caused by the Wi-Fi deauthentication attack on video quality (e.g., slow down, freeze, fast forward, low resolution) to the popular benchmark datasets for video anomaly detection. Our experiments with several state-of-the-art anomaly detection models 
show that the attackers can significantly undermine the reliability of video anomaly detection systems by causing frequent false alarms and hiding physical anomalies from the surveillance system. 

\end{abstract}

\section{Introduction}
\label{sec:intro}

In recent years, video anomaly detection has become 
a popular research topic with the increasing availability of video-recording hardware and the promising capabilities of artificial intelligence (AI) algorithms. 
The fast growing surveillance video streams prevent manual monitoring by human operators and necessitate automated monitoring by AI. Deep Neural Networks (DNNs), in particular Convolutional Neural Networks (CNNs), are known to provide promising results for video anomaly detection. 
Since 
video surveillance 
is often used for 
providing physical safety and security by detecting 
important events such as accidents, burglaries, or illegal activities, 
automated anomaly detection systems need to be dependable, secure, and robust to attacks. Because of the sensitive nature of 
video surveillance, raising frequent false alarms or not being able to detect the important anomalies could have 
serious consequences in real world. In this paper, we 
show that with an 
easy-to-implement
adversarial machine learning attack to 
state-of-the-art video
anomaly detection algorithms, it is possible to 
cause frequent 
false alarms or 
prevent 
anomalies from being detected.

Adversarial machine learning attacks 
damage
the liability of machine learning systems. Although originally adversarial attacks 
have been 
widely investigated for image recognition, lately, attacks 
targeting 
video action recognition models started to gain attention \cite{inkawhich2018adversarial, wei2019sparse, li2021adversarial, pony2021over}.
DNNs, in particular CNNs, which are commonly used in video understanding models,
are highly vulnerable to adversarial attacks. \cite{szegedy2013intriguing, goodfellow2014explaining}. The fundamental characteristic of an adversarial attack is to create input 
data
with perturbations which are undetectable by humans while they result in 
errors
for the machine learning algorithms. 
In the literature, adversarial data
are mostly generated by adding perturbations to the original inputs 
with an inherent assumption that the attacker somehow has access to the data input to the machine learning model. While this assumption easily holds for machine learning models open to public use or by assuming the attacker is an authorized user (or an intruder), it needs further justification for video surveillance systems which are typically private systems with restricted access. 
In this paper, we 
demonstrate 
how an adversarial machine learning attack can be performed against a video anomaly detection system through an easy-to-implement cyber-attack. 

Wi-Fi deauthentication attack is a type of denial-of-service
(DoS)
attack, 
which targets the communication between clients and a Wi-Fi access point. The main purpose of this attack is to send deauthentication packets to the Wi-Fi 
access point and originate a network outage for the connected devices as a result. In our experiments, we observed that if surveillance cameras are connected to a network which is being attacked by Wi-Fi deauthentication, then the transmitted videos will have malfunctions. These malfunctions include frame drops, lagging, slowing down, freezing, 
and drops in the quality of the frames. 
We demonstrate that
DNN-based anomaly detection methods 
are prone to increased false alarms due to such malfunction effects. 
Furthermore, 
we show that an attacker can use the freezing effect to hide a physical anomaly from the surveillance system by carefully synchronizing the cyber-attack with the physical anomaly. 

In summary, our goal in this paper is to demonstrate that 
an easy-to-implement 
cyber-attack can produce an adversarial machine learning attack 
to a video surveillance system.
To this end,
we followed 
the following steps:
(i) First, 
we created a testbed with a security camera connected to Wi-Fi, a router, and an attacker software to observe the effects of Wi-Fi deauthentication attack on the surveillance video.
(ii) Then, we applied these effects to two 
popular
benchmark datasets, namely 
the 
CUHK Avenue 
and ShanghaiTech Campus datasets, 
to obtain adversarial data for testing benchmark algorithms.
(iii) Finally, 
we used our new datasets 
simulating the Wi-Fi deauthentication attack to evaluate performance of several state-of-the-art DNN-based anomaly detection methods:
Future Frame Prediction \cite{liu2018future}, Memory-guided Normality for Anomaly Detection (MNAD) \cite{park2020learning}, Modular Online Video Anomaly Detector (MOVAD) \cite{doshi2022modular}. 
Our contributions in this paper 
can be summarized as follows:
\begin{itemize}
\item We propose the first adversarial machine learning attack that targets video anomaly detection models.
\item  We 
use
Wi-Fi deauthentication attack to generate 
an adversarial machine learning attack, and to our knowledge, this is the first approach which uses a cyber-attack for 
performing adversarial machine learning attack on 
video understanding models (e.g., action recognition, anomaly detection, etc.).
\end{itemize}

\begin{figure}[t]
  \centering
   \includegraphics[width=0.8\linewidth]{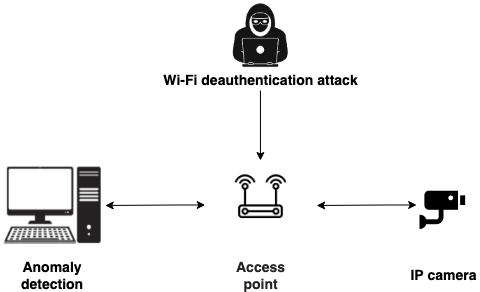}
   \caption{Digram of Wi-Fi deauthentication attack, 
   which targets the Wi-Fi connection between the surveillance camera and the anomaly detection model.}
   \label{fig:wifiem}
\end{figure}


\section{Related Works}

Video
anomaly detection is a fast-growing subject in machine learning. The main goal of 
video 
anomaly detection is successfully reporting 
unexpected events while controlling false alarms. 
Although the early methods 
use hand-designed features
\cite{chaudhry2009histograms, colque2016histograms}, most of the recent 
works use DNNs
\cite{hasan2016learning, hinami2017joint, luo2017revisit, ravanbakhsh2019training, sabokrou2018adversarially, xu2015learning,markovitz2020graph, ramachandra2020street, ionescu2019object,doshi2020continual,doshi2022rethinking}. 
The recent DNN-based methods can be categorized into two main groups, prediction-based \cite{liu2018future,lee2019bman,dong2020dual,doshi2021online} and reconstruction-based \cite{gong2019memorizing,hasan2016learning,luo2017revisit,nguyen2019anomaly,park2020learning}. They, respectively, compare the error in predicting the next frame and reconstructing the observed frames with a threshold to detect anomalies. 


Although adversarial attacks on images 
have been 
investigated for several years since 
\cite{goodfellow2014explaining}, the vulnerabilities of video 
recognition models 
only recently attracted attention \cite{jiang2019black,wei2020heuristic,yan2020sparse,zhang2020motion,li2021adversarial,pony2021over}.
Adversarial attacks on videos have been investigated in both white-box and black-box settings. The difference between these two settings is the accessibility to the 
knowledge of target model.
While in the white-box setting, an attacker has full access to the model including its architecture and parameters; in the black-box setting, the attacker has access only to the output labels.

Adversarial black-box attacks for video classification models are relatively new compared to the white-box attacks. The most common black-box attack pattern is sending queries to the target model and estimating gradients to generate adversarial data.  PatchAttack (V-BAD) \cite{jiang2019black} is the first proposed black-box video attack. They proposed a method to generate perturbations for each frame of a video, 
and
then 
updated their perturbations with queries. HeuristicAttack \cite{wei2020heuristic} is another black-box adversarial attack which uses a query-based attack strategy to heuristically select 
the key frames 
to be attacked. Similarly, SparseAttack \cite{yan2020sparse} aims to select key frames by using reinforcement learning. Motion-sampler attack \cite{zhang2020motion} is another attack which uses query-based strategy, but they aim to take advantage of using motion-excited sampler to reduce required queries. Geo-Trap \cite{li2021adversarial} uses geometric transformations to parameterize the temporal space, and as a result they generate successful perturbations with fewer queries. 

The attack we introduce in this paper is also a black-box attack, but it differs from 
the existing black-box attacks in two main aspects.
The first difference is that all of these attacks 
target action recognition models while our attack targets anomaly detection models. And the other difference is that while these attacks use query-based strategies to 
shape their perturbations, we use a cyber-attack to generate our adversarial data.

Over-the-Air Flickering Attack \cite{pony2021over} is a white-box adversarial machine learning attack. 
Similar to our work, they demonstrate how to implement their attack in a real-world scenario. 
Their idea is attacking the RGB stream of the data with the help of RGB bulbs. Therefore, in terms of 
applicability in real-world, it is the most similar attack to ours. However, their target models are action recognition 
models and they use a physical intervention, instead of a cyber-attack, to generate adversarial data. 
\section{Methodology}


Existing video anomaly detection 
methods typically utilize appearance and/or motion features.
Specifically, 
appearance features 
are extracted using object detection models, and
motion 
features are extracted through optical flow computation. 
However, the accuracy of object detection and optical flow computation are conditional on the video being transmitted without any disruptions and signal loss. 
Moreover, it is known that wireless networks are far more vulnerable to cyber-attacks than wired networks. \cite{bogdanoski2017analysis}. Due to the susceptible nature of wireless networks, we believe that an automated video surveillance framework would be adversely affected if its associated wireless network was attacked, especially in a stealthy manner. Instead of a brute-force DoS attack to completely stop the video streaming, which would make the attack obvious to the defender, attackers may balance the intensity of their disruption to make it look like a natural connectivity issue. 

To evaluate our hypothesis, we set up a testbed using an Amazon Blink IP camera transmitting data to a mobile app, emulating a video surveillance system, as shown in Fig. \ref{fig:wifiem}. We targeted our system via a Wi-Fi deauthentication attack 
and an ARP spoofing attack using the Evil Limiter software\footnote{\url{https://github.com/bitbrute/evillimiter}}. By sending deauthentication packets, an attacker can potentially target the communication between clients and routers, which can result in clients experiencing connection outages. On the other hand, Evil Limiter limits the bandwidth of devices on a local network, effectively disrupting communication. Using an ARP spoofing tool like Evil Limiter allows the adversary more control over the attack magnitude since the goal of the adversary is to stealthily attack a video surveillance framework. 

Next, we discuss the threat model posed by the proposed wireless network attacks, as well as a simulated data injection attack. Our goal in the following sections is to present the observed effects of the considered attacks to the video stream and demonstrate the effect of adversarial data following these effects on the existing state-of-the-art anomaly detection methods.

\subsection{Threat Model}
\label{sec:effects}
\textbf{Cyber-Attacks to Wi-Fi Network:} We begin by exploring the effects of the two network attacks on the the transmitted video. First, we limited the bandwidth to mimic a scenario where an adversary attempts to stealthily disrupt the video stream by affecting the communication quality. Then, we attempted a more aggressive attack using deauthentication packets, which leads to the security camera disconnecting from the Wi-Fi node. After empirically analyzing the received video under both attacks, we broadly categorize the observed effects into four classes:   


\begin{itemize}
\item \emph{Slowing effect:} Once the available bandwidth for the security camera drops below a certain threshold, we begin to notice the transmitted video being sluggish or playing in slow motion, which is caused due to a longer delay in frames being transmitted from the memory buffer.

\item \emph{Low-resolution effect:} During the attack, we also notice that the camera begins to transmit video at a lower resolution as a way of compensating for poor network quality.

\item \emph{Freezing effect:} In the case of a deauthentication attack, the camera is unable to transmit any more frames, and hence the video feed appears to be frozen. Empirically, the duration of the freeze is related to the intensity of the attack.




\item \emph{Fast-forwarding effect:} Finally, after the connection gets restored, the video fast forwards to the current frame, causing a rapid motion. Usually, the fast forward effect lasts for a fixed duration, irrespective of how long the attack lasted.

\end{itemize}



\textbf{Data Injection Attack:} In addition to the proposed cyber-attacks, we also consider an imaginary scenario in which the attacker has access to the transmitted data and thus is able to perform a data injection attack. Here, the adversary infrequently freezes or replicates the frames for an extremely small duration, without causing any noticeable difference to the human eye.

Theoretically, it is difficult to detect a network attack since it is almost indistinguishable from video corruption due to ordinary connection problems. It is important to note that while slowing, freezing, and fast forwarding occur consequently, low resolution can be observed during the entire attack duration. On the other hand, the video transmitted during a replicating attack seems identical to the original video and cannot be differentiated by even a human operator. To evaluate the effectiveness of existing state-of-the-art approaches, we imitate these effects on public benchmark datasets and generate a new set of adversarial videos, as discussed next in Section \ref{sec:datasets}. 



\subsection{Generating Adversarial Datasets}
\label{sec:datasets}



We leverage two popular video anomaly detection datasets, the CUHK Avenue dataset \cite{lu2013abnormal} and the ShanghaiTech Campus dataset \cite{luo2017revisit}. We imitated the effects discussed in Section \ref{sec:effects} and generated new adversarial datasets.
Specifically, we consider 
two
parameters for generating a new adversarial video: 
the duration $D$ of the attack, and the attack onset time $t$. 
The attack onset time $t$ is randomly selected for each video. We next describe each generated dataset in detail. 




\begin{itemize}
\item \textbf{Slow-Freeze-Fast Dataset:} Similar to what we observe during a network attack, we first introduce a slowing effect followed by the freezing and fast-forwarding effects. The lengths of these portions are 
respectively given by $D/3$, $D/6$, and $D/2$.


\item \textbf{Low-Resolution Dataset:} For the low-resolution dataset, 
we lower the resolution of all frames 
for a duration of $D$.


\item \textbf{Combined Dataset:} Finally, we also generate a combined dataset by combining the effects from the first two datasets.
\end{itemize}



\section{Experiments}
\label{sec:experiments}

In this section, we evaluate the performance of existing state-of-the-art approaches, namely the Future Frame Prediction (FFPN) \cite{liu2018future}, Memory-guided Normality for Anomaly Detection (MNAD) \cite{park2020learning}, and Modular Online Video Anomaly Detector (MOVAD) \cite{doshi2022modular} on the generated adversarial datasets. We specifically consider these approaches since their implementations are publicly available, and can be easily modified to work with the proposed datasets. We evaluate the performance of all existing approaches using the commonly used Area under Curve (AUC) metric.

\begin{figure}[tbh]
  \centering
   \includegraphics[width=\linewidth]{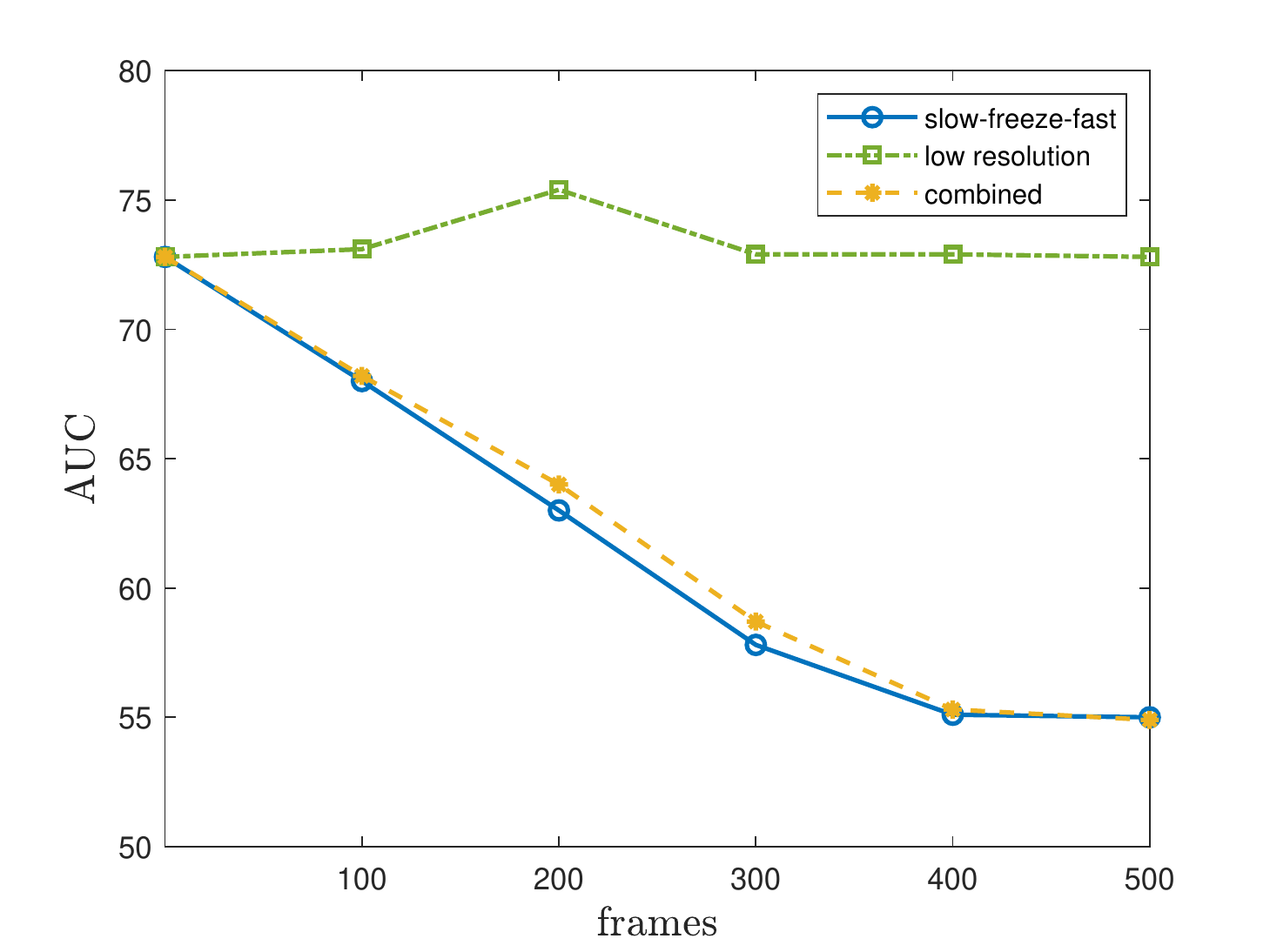}
   \includegraphics[width=\linewidth]{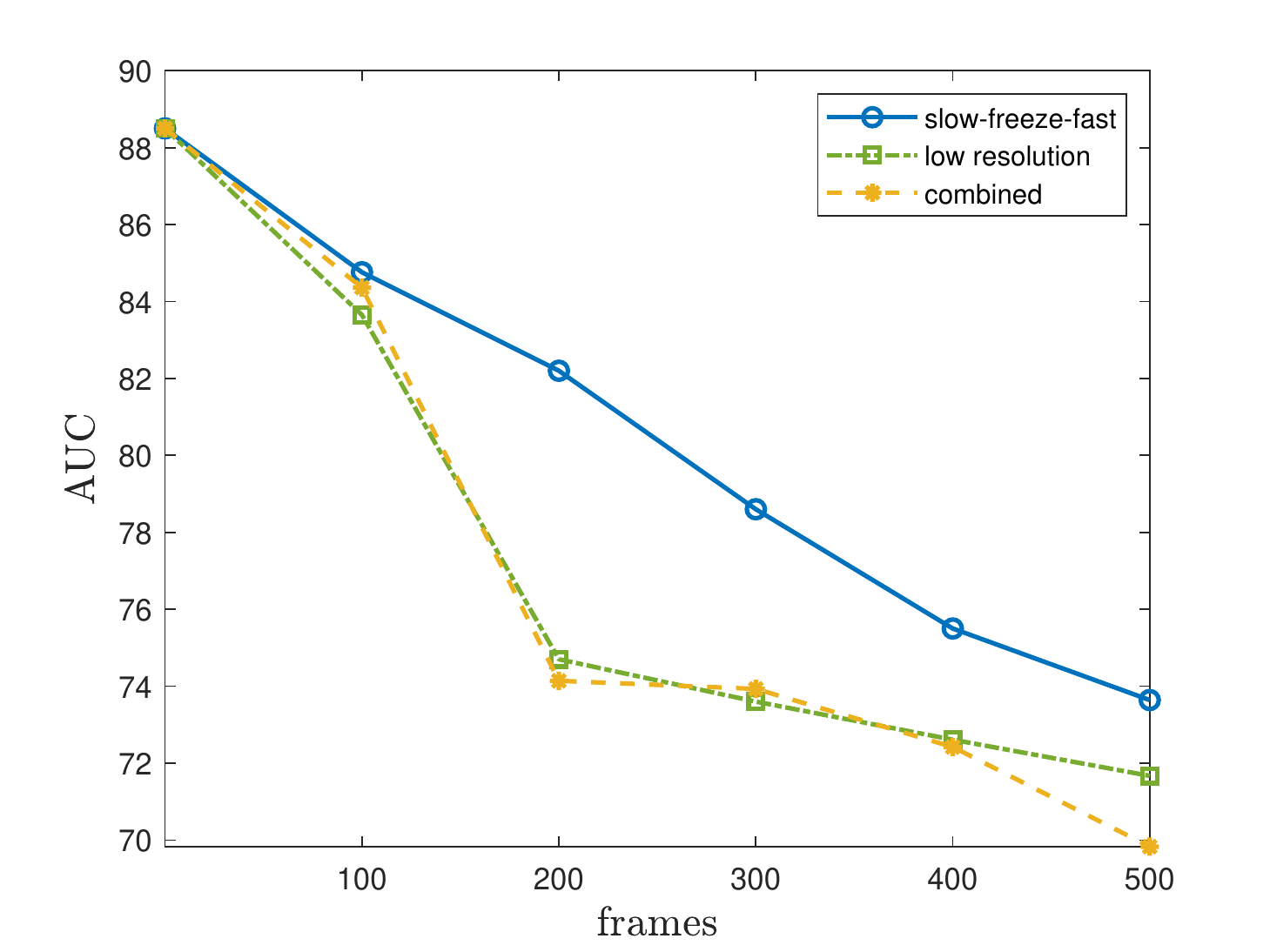}
   \includegraphics[width=\linewidth]{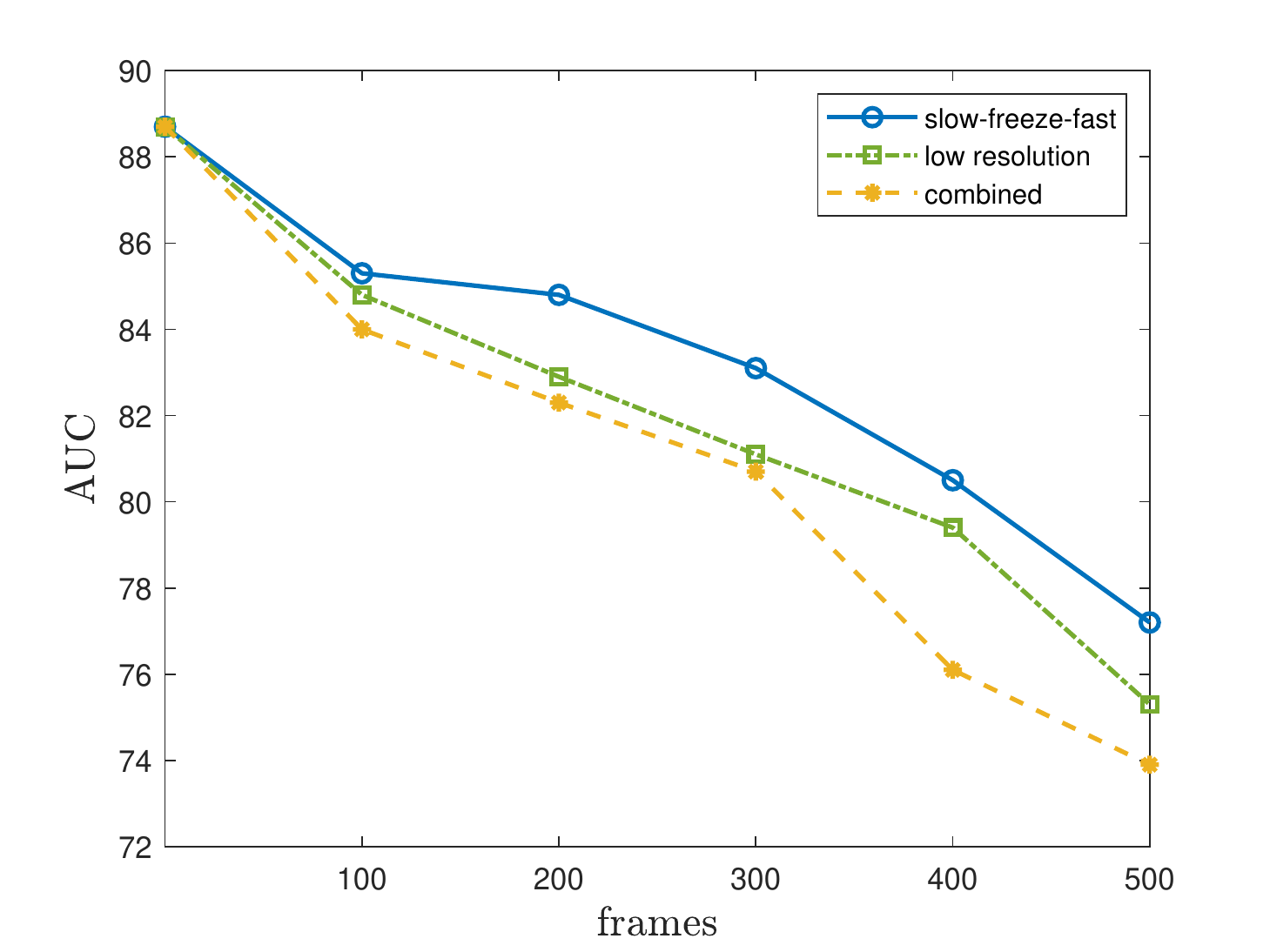}
   \caption{
   Performance drops of FFPN \cite{liu2018future} with adversarial ShanghaiTech dataset (top), 
   MNAD \cite{park2020learning} with adversarial Avenue dataset (middle), 
   MOVAD \cite{doshi2022modular} with adversarial Avenue dataset (bottom). }
   \label{fig:slowfast}
\end{figure}


\subsection{Network Attacks}

\begin{table}
  \centering
  \begin{tabular}{|c|c|c|}
    \hline
    Method & Avenue $\to$ SFF & ShanghaiTech $\to$ SFF \\
    \hline
    FFPN \cite{liu2018future} & 0.85 $\to$ 0.684 & 0.728 $\to$ 0.55 \\
    MNAD \cite{park2020learning} & 0.885 $\to$ 0.74 & - \\
    MOVAD \cite{doshi2022modular} & 0.887 $\to$ 0.77 & - \\
    \hline
  \end{tabular}
  \caption{Performance (AUC) drop in FFPN \cite{liu2018future}, MNAD \cite{park2020learning}, and MOVAD \cite{doshi2022modular} due to the Slow-Fast-Freeze effect. The attack duration is $D=500$.}
  \label{tab:slowfast}
\end{table}

\begin{figure*}[t]
\centering
   \includegraphics[width=0.7\linewidth]{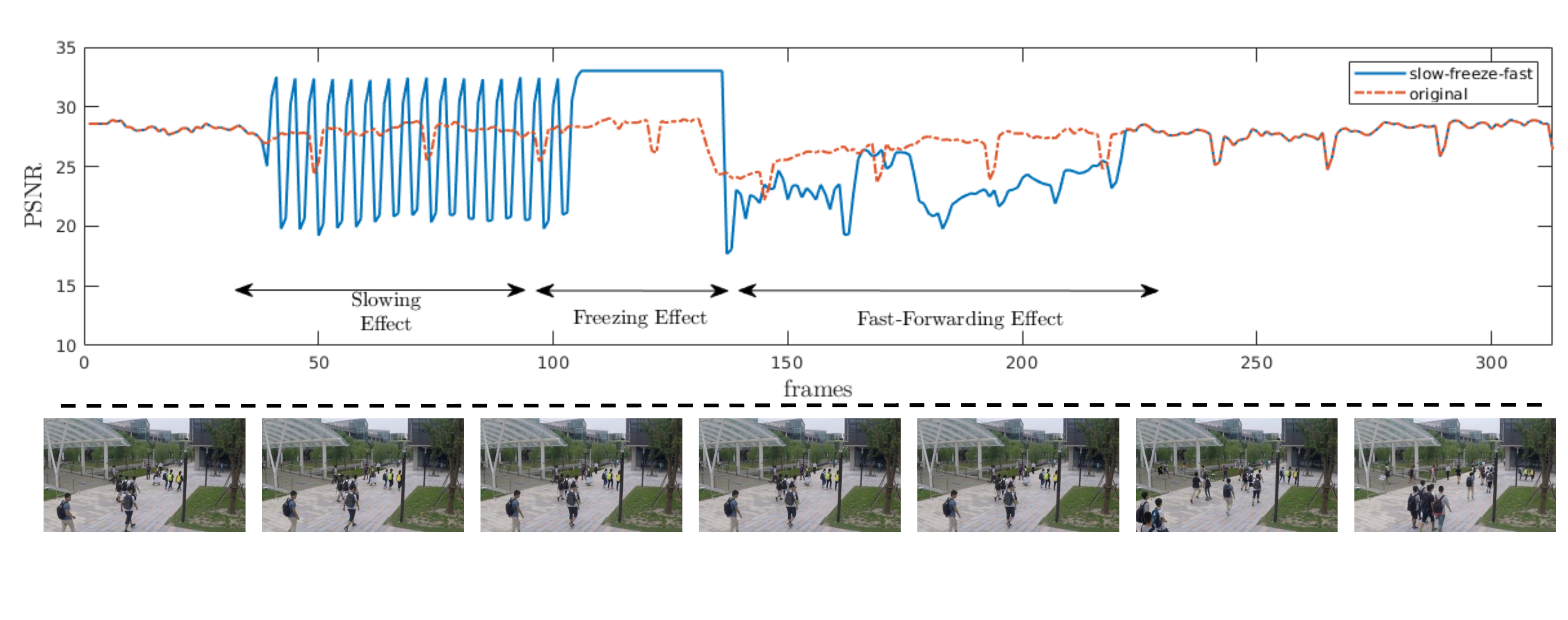}\\
   \includegraphics[width=0.7\linewidth]{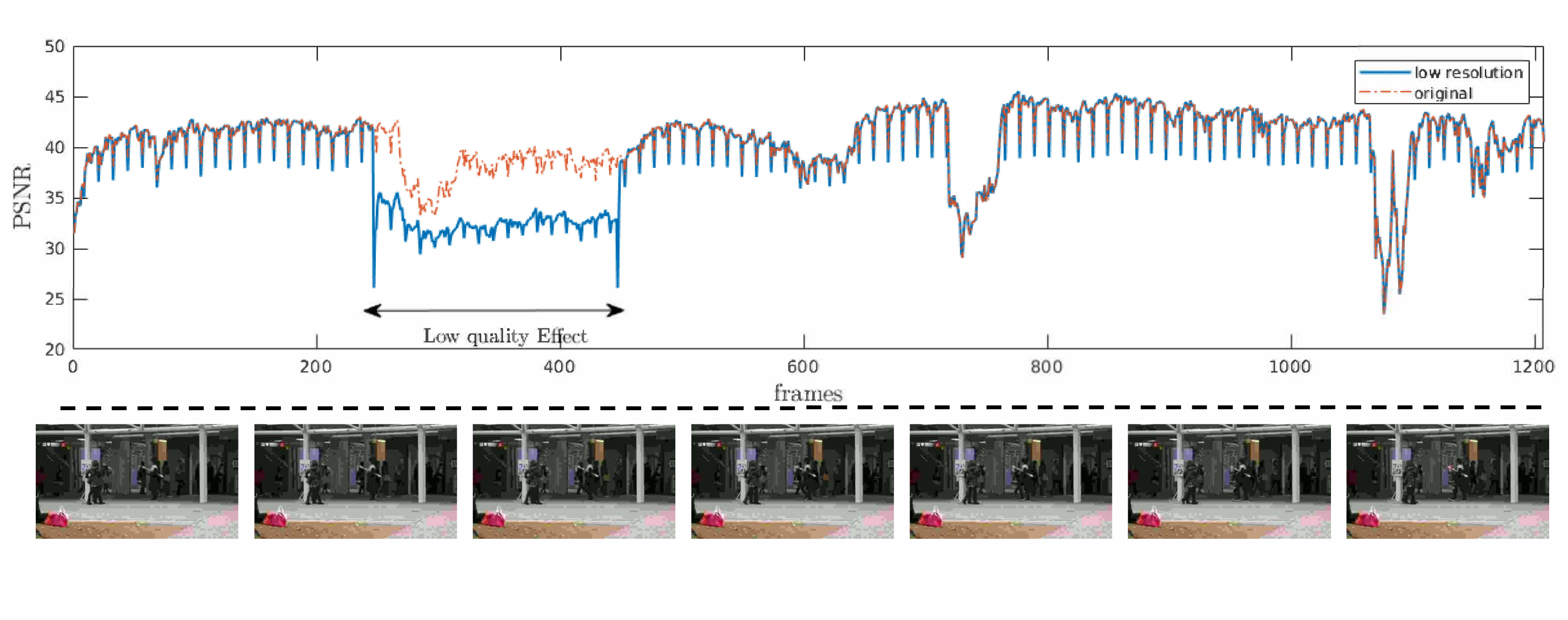}
   \caption{
   The variation in PSNR statistic based on the effect caused by the adversarial attack. The first figure represents the statistic for the ShanghaiTech dataset under the slowing, freezing, and fast-forwarding effects. The second figure represents the Avenue dataset under the low-resolution effect.}
   \label{fig:PSNR_effect}
\end{figure*}

Since all the 
considered 
approaches leverage motion features for detecting anomalies
like the vast majority of existing methods, we notice a steady drop in their performances as the attack intensity increases. 
In Fig. \ref{fig:slowfast}, we compare the performance of the considered methods on the benchmark datasets by varying the attack duration $D$. It should be noted that an attack lasting for 500 frames is equivalent to 17 seconds, which is a reasonably short duration. 
As summarized in Table \ref{tab:slowfast}, the considered state-of-the-art methods are all prone to the Slow-Freeze-Fast effect. It is interesting to see that FFPN is robust to the low-resolution effect (Fig. \ref{fig:slowfast}). 
The trained models for MNAD and MOVAD on ShanghaiTech dataset are not publicly available. 



\subsection{Increased False Alarms}

For detecting anomalies, both MNAD and FFPN compute the Peak Signal to Noise Ratio (PSNR) between the actual frame and predicted frame. Generally, a drop in the PSNR values constitute an anomaly. Hence, apart from the AUC-based performance metrics, we also observe the adversarial effect of different network attacks on the PSNR values in Fig. \ref{fig:PSNR_effect}. As discussed in Section \ref{sec:effects}, there are four distinct types of effects caused by network attacks: slowing, freezing, fast forwarding, and low resolution. Each of these effects has a unique impact on the PSNR values. Slow effect results in frequent drops and rises; fast-forwarding effect results in a large initial drop and maintained the PSNR value at a low level throughout the effect; low resolution effect results in drops in PSNR values relative to the original values; and finally, freezing effect results in stabilized PSNR values. 
With the exception of freezing, all effects result in a decrease in PSNR values, leading to false alarms.



\begin{figure*}[t]
  \minipage{0.33\textwidth}
   \includegraphics[width=1.1\linewidth]{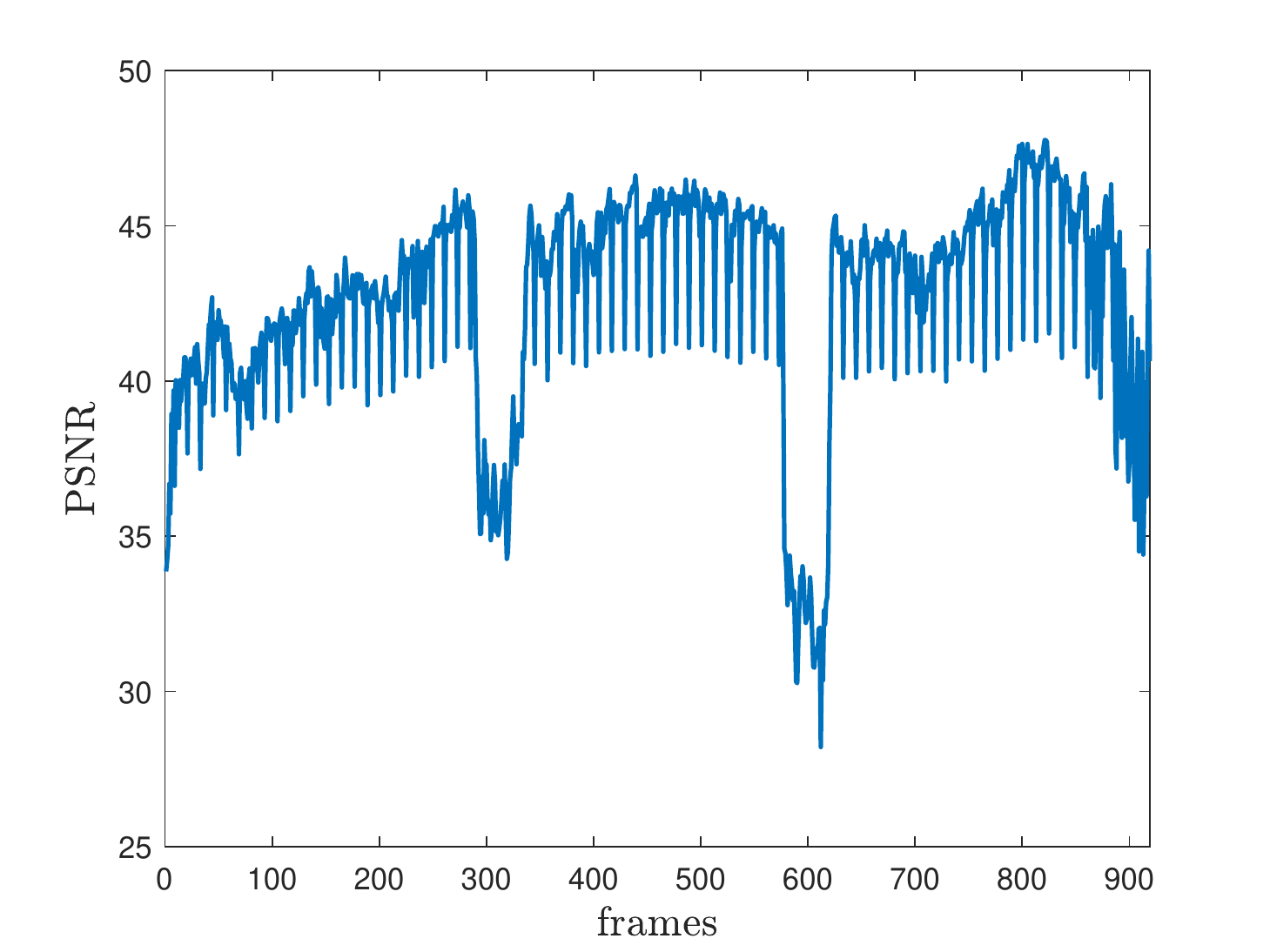}
   \endminipage \hfill
   \minipage{0.33\textwidth}
   \includegraphics[width=1.1\linewidth]{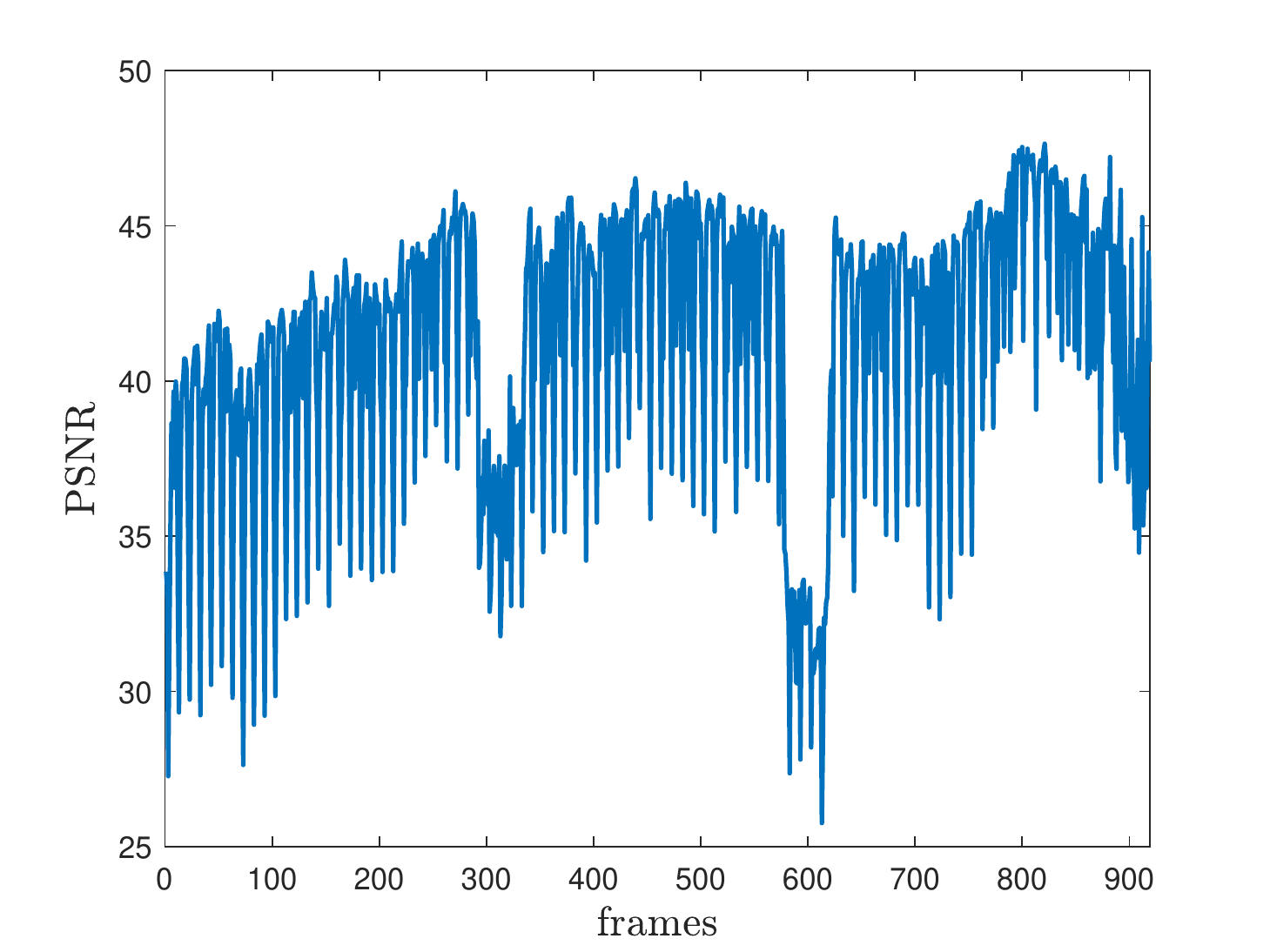}
   \endminipage \hfill
   \minipage{0.33\textwidth}
   \includegraphics[width=1.1\linewidth]{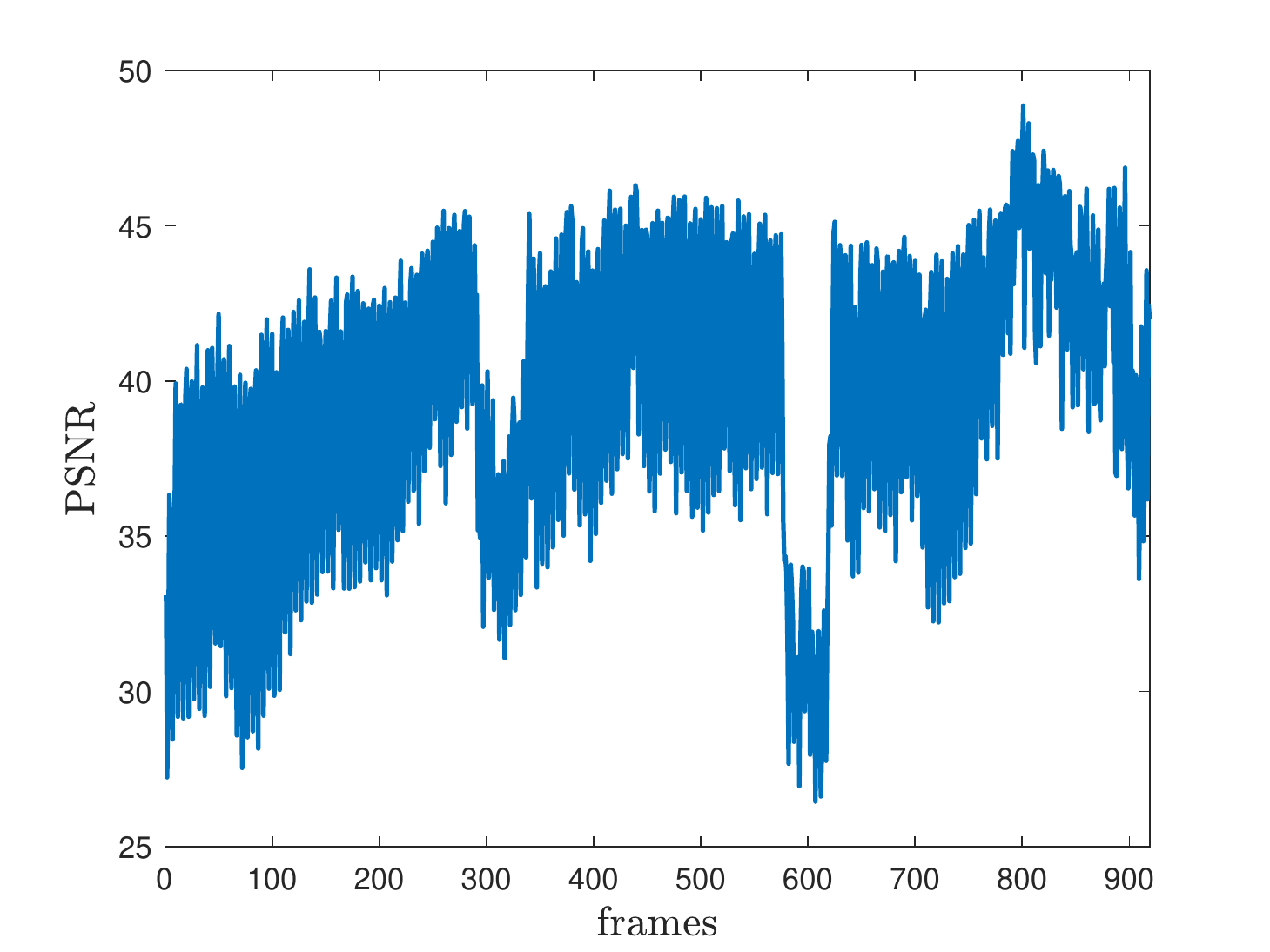}
   \endminipage
   \caption{Comparison of the PSNR statistic based on the magnitude of the replicating attack on the third test video from the Avenue dataset. We show the PSNR statistic for the original video (left), the PSNR statistic on Replicated Dataset 1 (middle) and Replicated Dataset 2 (right).}
   \label{fig:replicated}
\end{figure*}

\subsection{Hiding a Physical Anomaly}

While slowing, fast-forwarding, and low-resolution effects result in false alarms, the freeze effect prevents the PSNR from dropping and thereby preventing an anomaly from being detected. 
From the empirical evidence using our testbed, we observe that the duration of the freeze effect is dependent on the intensity of the Wi-Fi deauthentication attack. Thus, if an attacker manages to synchronize a cyber-attack with a physical attack, such as burglary, it is possible to conceal the physical attack until the freeze attack is completed. 

\begin{figure}[t]
  \centering
   \includegraphics[width=\linewidth]{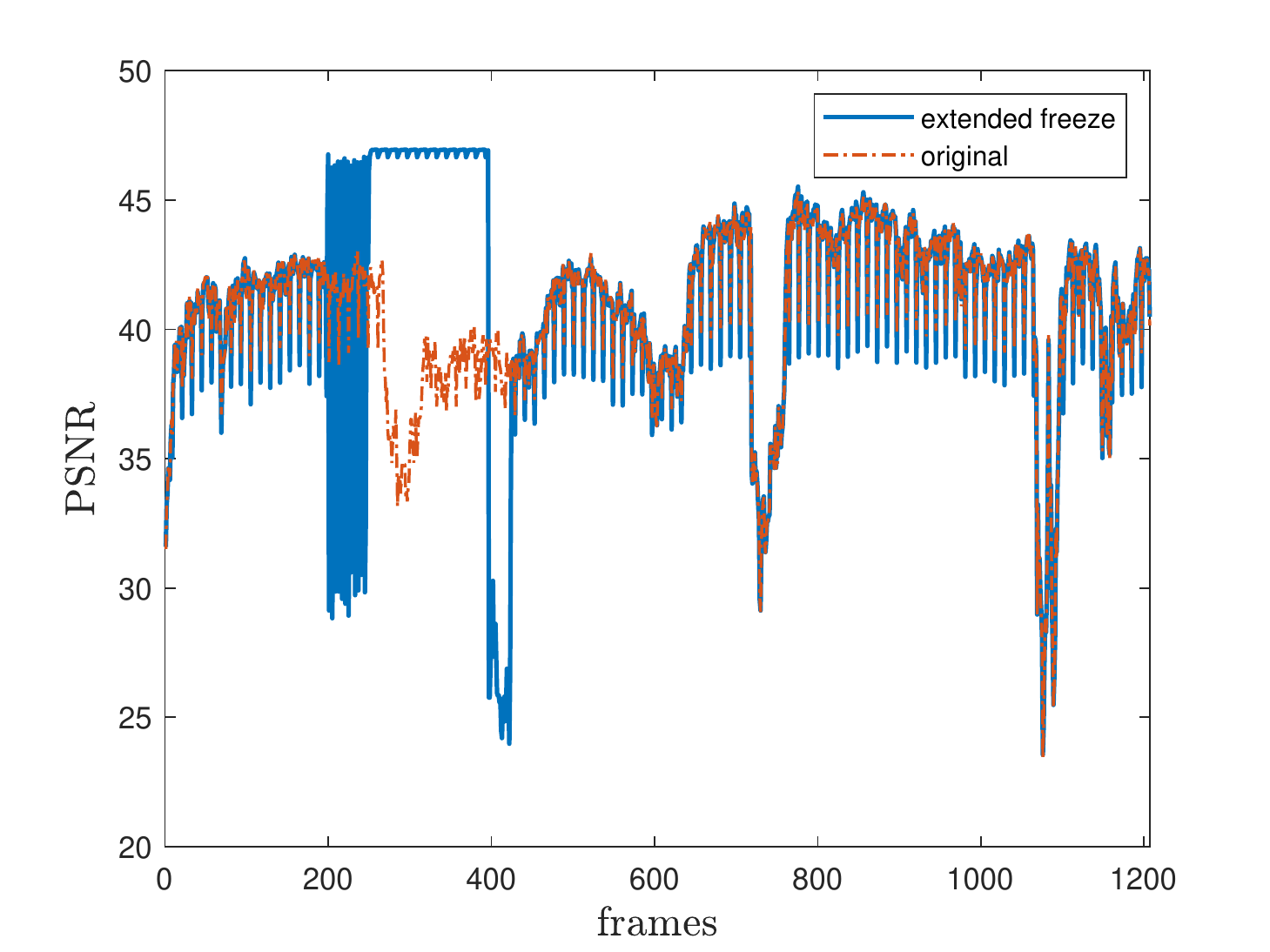}
   \caption{PSNR values of a video with extended freeze effect compared to the original video. The drop in PSNR, which corresponds to a physical anomaly successfully detected in the original video, is masked by the freeze effect.}
   \label{fig:extended_freeze}
\end{figure}

To validate our hypothesis, we created a new adversarial video using the Avenue dataset's second test video. To generate this adversarial video, we followed the same procedure as described in Section \ref{sec:datasets} for generating the Slow-Freeze-Fast dataset. However, we 
increased the duration of the freeze effect such that it is three times longer than the duration of the slow effect. 
We applied these effects to the section that contains the anomaly. We show the effect of extended freeze attack in Fig. \ref{fig:extended_freeze}, where we notice that there is no change in the PSNR statistic when the anomaly occurs. 




\subsection{Replicating Attack}

\begin{table}
  \centering
  \begin{tabular}{|c|c|c|}
    \hline
    Dataset & MNAD \cite{park2020learning} & FFPN \cite{liu2018future} \\
    \hline
    Avenue & 0.885 & 0.85 \\
    Replicated 1 & 0.831 & 0.81\\
    Replicated 2 & 0.819 & 0.795\\
    \hline
  \end{tabular}
  \caption{Performance (AUC) drop in MNAD and FFPN due to Replicating Attack.}
  \label{tab:rep}
\end{table}




We also consider an imaginary scenario in which the adversary has access to the surveillance system, and hence can directly manipulate the data being transmitted, similar to the scenario commonly considered in the adversarial machine learning literature. The goal of the adversary would be to raise unnecessary false alarms in a stealthy manner. Specifically, we consider a replicating attack in which the adversary replicates or freezes the frames for an extremely small duration such that it is unnoticeable to the human eye. To test the effectiveness of our method, we generate two datasets from the Avenue dataset. In the first dataset, we replicate one frame every 10 frames. Similarly, for the second dataset we replicate two frames every 10 frames. We notice that the AUC drops from 0.855 to 0.795 for the FFPN approach and from 0.885 to 0.819 for the MNAD model. We show the performance drops caused by the replicating attack in Table \ref{tab:rep}.

More importantly, we observe many frequent drops in the PSNR values, which indicate that the system may raise false alarms constantly. In Fig. \ref{fig:replicated}, we show the PSNR statistic of the third video from the Avenue dataset, for both the original video and videos from our replicating attack datasets.



\section{Conclusions}

Adversarial machine learning attacks on video recognition models have recently become a subject of increasing interest. In this work, we, for the first time in the literature, demonstrated an adversarial attack directed at video anomaly detection methods.
Our adversarial attacks are generated using easy-to-implement cyber-attacks, in particular the Wi-Fi deauthentication and ARP spoofing attacks.
We observe significant performance deterioration in three state-of-the-art anomaly detection methods because of our attack. Additionally, upon examining their decision statistics, we discovered numerous false alarms caused by our network attacks. We also demonstrate that with the appropriate synchronization, it is possible to conceal a real-world anomaly such as a burglary using the explained cyber-attack. 
For future work, we intend to conduct additional research on adversarial attacks on video anomaly detection systems and develop defense mechanisms against them.



{\small
\bibliographystyle{ieee_fullname}
\bibliography{egbib}

\begin{thebibliography}{10}\itemsep=-1pt

\bibitem{bogdanoski2017analysis}
Mitko Bogdanoski, Pero Latkoski, and Aleksandar Risteski.
\newblock Analysis of the impact of authrf and assrf attacks on ieee 802.11
  e-based access point.
\newblock {\em Mobile Networks and Applications}, 22(5):834--843, 2017.

\bibitem{chaudhry2009histograms}
Rizwan Chaudhry, Avinash Ravichandran, Gregory Hager, and Ren{\'e} Vidal.
\newblock Histograms of oriented optical flow and binet-cauchy kernels on
  nonlinear dynamical systems for the recognition of human actions.
\newblock In {\em 2009 IEEE Conference on Computer Vision and Pattern
  Recognition}, pages 1932--1939. IEEE, 2009.

\bibitem{colque2016histograms}
Rensso Victor Hugo~Mora Colque, Carlos Caetano, Matheus Toledo~Lustosa de
  Andrade, and William~Robson Schwartz.
\newblock Histograms of optical flow orientation and magnitude and entropy to
  detect anomalous events in videos.
\newblock {\em IEEE Transactions on Circuits and Systems for Video Technology},
  27(3):673--682, 2016.

\bibitem{dong2020dual}
Fei Dong, Yu Zhang, and Xiushan Nie.
\newblock Dual discriminator generative adversarial network for video anomaly
  detection.
\newblock {\em IEEE Access}, 8:88170--88176, 2020.

\bibitem{doshi2020continual}
Keval Doshi and Yasin Yilmaz.
\newblock Continual learning for anomaly detection in surveillance videos.
\newblock In {\em Proceedings of the IEEE/CVF conference on computer vision and
  pattern recognition workshops}, pages 254--255, 2020.

\bibitem{doshi2021online}
Keval Doshi and Yasin Yilmaz.
\newblock Online anomaly detection in surveillance videos with asymptotic bound
  on false alarm rate.
\newblock {\em Pattern Recognition}, 114:107865, 2021.

\bibitem{doshi2022modular}
Keval Doshi and Yasin Yilmaz.
\newblock A modular and unified framework for detecting and localizing video
  anomalies.
\newblock In {\em Proceedings of the IEEE/CVF Winter Conference on Applications
  of Computer Vision}, pages 3982--3991, 2022.

\bibitem{doshi2022rethinking}
Keval Doshi and Yasin Yilmaz.
\newblock Rethinking video anomaly detection-a continual learning approach.
\newblock In {\em Proceedings of the IEEE/CVF Winter Conference on Applications
  of Computer Vision}, pages 3961--3970, 2022.

\bibitem{gong2019memorizing}
Dong Gong, Lingqiao Liu, Vuong Le, Budhaditya Saha, Moussa~Reda Mansour, Svetha
  Venkatesh, and Anton van~den Hengel.
\newblock Memorizing normality to detect anomaly: Memory-augmented deep
  autoencoder for unsupervised anomaly detection.
\newblock In {\em Proceedings of the IEEE/CVF International Conference on
  Computer Vision}, pages 1705--1714, 2019.

\bibitem{goodfellow2014explaining}
Ian~J Goodfellow, Jonathon Shlens, and Christian Szegedy.
\newblock Explaining and harnessing adversarial examples.
\newblock {\em arXiv preprint arXiv:1412.6572}, 2014.

\bibitem{hasan2016learning}
Mahmudul Hasan, Jonghyun Choi, Jan Neumann, Amit~K Roy-Chowdhury, and Larry~S
  Davis.
\newblock Learning temporal regularity in video sequences.
\newblock In {\em Proceedings of the IEEE conference on computer vision and
  pattern recognition}, pages 733--742, 2016.

\bibitem{hinami2017joint}
Ryota Hinami, Tao Mei, and Shin'ichi Satoh.
\newblock Joint detection and recounting of abnormal events by learning deep
  generic knowledge.
\newblock In {\em Proceedings of the IEEE International Conference on Computer
  Vision}, pages 3619--3627, 2017.

\bibitem{inkawhich2018adversarial}
Nathan Inkawhich, Matthew Inkawhich, Yiran Chen, and Hai Li.
\newblock Adversarial attacks for optical flow-based action recognition
  classifiers.
\newblock {\em arXiv preprint arXiv:1811.11875}, 2018.

\bibitem{ionescu2019object}
Radu~Tudor Ionescu, Fahad~Shahbaz Khan, Mariana-Iuliana Georgescu, and Ling
  Shao.
\newblock Object-centric auto-encoders and dummy anomalies for abnormal event
  detection in video.
\newblock In {\em Proceedings of the IEEE Conference on Computer Vision and
  Pattern Recognition}, pages 7842--7851, 2019.

\bibitem{jiang2019black}
Linxi Jiang, Xingjun Ma, Shaoxiang Chen, James Bailey, and Yu-Gang Jiang.
\newblock Black-box adversarial attacks on video recognition models.
\newblock In {\em Proceedings of the 27th ACM International Conference on
  Multimedia}, pages 864--872, 2019.

\bibitem{lee2019bman}
Sangmin Lee, Hak~Gu Kim, and Yong~Man Ro.
\newblock Bman: bidirectional multi-scale aggregation networks for abnormal
  event detection.
\newblock {\em IEEE Transactions on Image Processing}, 29:2395--2408, 2019.

\bibitem{li2021adversarial}
Shasha Li, Abhishek Aich, Shitong Zhu, Salman Asif, Chengyu Song, Amit
  Roy-Chowdhury, and Srikanth Krishnamurthy.
\newblock Adversarial attacks on black box video classifiers: Leveraging the
  power of geometric transformations.
\newblock {\em Advances in Neural Information Processing Systems}, 34, 2021.

\bibitem{liu2018future}
Wen Liu, Weixin Luo, Dongze Lian, and Shenghua Gao.
\newblock Future frame prediction for anomaly detection--a new baseline.
\newblock In {\em Proceedings of the IEEE conference on computer vision and
  pattern recognition}, pages 6536--6545, 2018.

\bibitem{lu2013abnormal}
Cewu Lu, Jianping Shi, and Jiaya Jia.
\newblock Abnormal event detection at 150 fps in matlab.
\newblock In {\em Proceedings of the IEEE international conference on computer
  vision}, pages 2720--2727, 2013.

\bibitem{luo2017revisit}
Weixin Luo, Wen Liu, and Shenghua Gao.
\newblock A revisit of sparse coding based anomaly detection in stacked rnn
  framework.
\newblock In {\em Proceedings of the IEEE international conference on computer
  vision}, pages 341--349, 2017.

\bibitem{markovitz2020graph}
Amir Markovitz, Gilad Sharir, Itamar Friedman, Lihi Zelnik-Manor, and Shai
  Avidan.
\newblock Graph embedded pose clustering for anomaly detection.
\newblock In {\em Proceedings of the IEEE/CVF Conference on Computer Vision and
  Pattern Recognition}, pages 10539--10547, 2020.

\bibitem{nguyen2019anomaly}
Trong-Nguyen Nguyen and Jean Meunier.
\newblock Anomaly detection in video sequence with appearance-motion
  correspondence.
\newblock In {\em Proceedings of the IEEE/CVF international conference on
  computer vision}, pages 1273--1283, 2019.

\bibitem{park2020learning}
Hyunjong Park, Jongyoun Noh, and Bumsub Ham.
\newblock Learning memory-guided normality for anomaly detection.
\newblock In {\em Proceedings of the IEEE/CVF Conference on Computer Vision and
  Pattern Recognition}, pages 14372--14381, 2020.

\bibitem{pony2021over}
Roi Pony, Itay Naeh, and Shie Mannor.
\newblock Over-the-air adversarial flickering attacks against video recognition
  networks.
\newblock In {\em Proceedings of the IEEE/CVF Conference on Computer Vision and
  Pattern Recognition}, pages 515--524, 2021.

\bibitem{ramachandra2020street}
Bharathkumar Ramachandra and Michael Jones.
\newblock Street scene: A new dataset and evaluation protocol for video anomaly
  detection.
\newblock In {\em The IEEE Winter Conference on Applications of Computer
  Vision}, pages 2569--2578, 2020.

\bibitem{ravanbakhsh2019training}
Mahdyar Ravanbakhsh, Enver Sangineto, Moin Nabi, and Nicu Sebe.
\newblock Training adversarial discriminators for cross-channel abnormal event
  detection in crowds.
\newblock In {\em 2019 IEEE Winter Conference on Applications of Computer
  Vision (WACV)}, pages 1896--1904. IEEE, 2019.

\bibitem{sabokrou2018adversarially}
Mohammad Sabokrou, Mohammad Khalooei, Mahmood Fathy, and Ehsan Adeli.
\newblock Adversarially learned one-class classifier for novelty detection.
\newblock In {\em Proceedings of the IEEE Conference on Computer Vision and
  Pattern Recognition}, pages 3379--3388, 2018.

\bibitem{szegedy2013intriguing}
Christian Szegedy, Wojciech Zaremba, Ilya Sutskever, Joan Bruna, Dumitru Erhan,
  Ian Goodfellow, and Rob Fergus.
\newblock Intriguing properties of neural networks.
\newblock {\em arXiv preprint arXiv:1312.6199}, 2013.

\bibitem{wei2019sparse}
Xingxing Wei, Jun Zhu, Sha Yuan, and Hang Su.
\newblock Sparse adversarial perturbations for videos.
\newblock In {\em Proceedings of the AAAI Conference on Artificial
  Intelligence}, volume~33, pages 8973--8980, 2019.

\bibitem{wei2020heuristic}
Zhipeng Wei, Jingjing Chen, Xingxing Wei, Linxi Jiang, Tat-Seng Chua, Fengfeng
  Zhou, and Yu-Gang Jiang.
\newblock Heuristic black-box adversarial attacks on video recognition models.
\newblock In {\em Proceedings of the AAAI Conference on Artificial
  Intelligence}, volume~34, pages 12338--12345, 2020.

\bibitem{xu2015learning}
Dan Xu, Elisa Ricci, Yan Yan, Jingkuan Song, and Nicu Sebe.
\newblock Learning deep representations of appearance and motion for anomalous
  event detection.
\newblock {\em arXiv preprint arXiv:1510.01553}, 2015.

\bibitem{yan2020sparse}
Huanqian Yan, Xingxing Wei, and Bo Li.
\newblock Sparse black-box video attack with reinforcement learning.
\newblock {\em arXiv preprint arXiv:2001.03754}, 2020.

\bibitem{zhang2020motion}
Hu Zhang, Linchao Zhu, Yi Zhu, and Yi Yang.
\newblock Motion-excited sampler: Video adversarial attack with sparked prior.
\newblock In {\em European Conference on Computer Vision}, pages 240--256.
  Springer, 2020.

\end{thebibliography}
}

\end{document}